\documentclass[10pt,twocolumn,letterpaper]{article}

\usepackage{iccv}
\usepackage{times}
\usepackage{epsfig}
\usepackage{graphicx}
\usepackage{amsmath}
\usepackage{amssymb}

\usepackage{booktabs}
\usepackage{float}
\usepackage{stfloats}
\usepackage{makecell}
\usepackage{multirow}
\usepackage{pifont}
\usepackage{soul}
\usepackage{color}
\usepackage{xcolor}
\usepackage{blindtext}
\usepackage{caption}
\usepackage[accsupp]{axessibility}  

\usepackage[pagebackref=true,breaklinks=true,letterpaper=true,colorlinks,bookmarks=false]{hyperref}

\iccvfinalcopy 

\def\iccvPaperID{5961} 

\ificcvfinal\fi

\begin{document}

\title{Isomer: Isomerous Transformer for Zero-shot Video Object Segmentation}

\author{
Yichen Yuan$^{1}$
\quad Yifan Wang$^{1*}$
\quad Lijun Wang$^{1}$
\quad Xiaoqi Zhao$^{1}$
\quad Huchuan Lu$^{1}$ \\
Yu Wang$^{2}$
\quad Weibo Su$^{2}$
\quad Lei Zhang$^{2,3}$ \\
$^{1}$School of Information and Communication Engineering, Dalian University of Technology, China \\
$^{2}$OPPO Research Institute
\quad $^{3}$The Hong Kong Polytechnic University
}

\twocolumn[{
\maketitle
\begin{center}
    \centering
    \vspace{-20pt}
    \includegraphics[width=18cm]{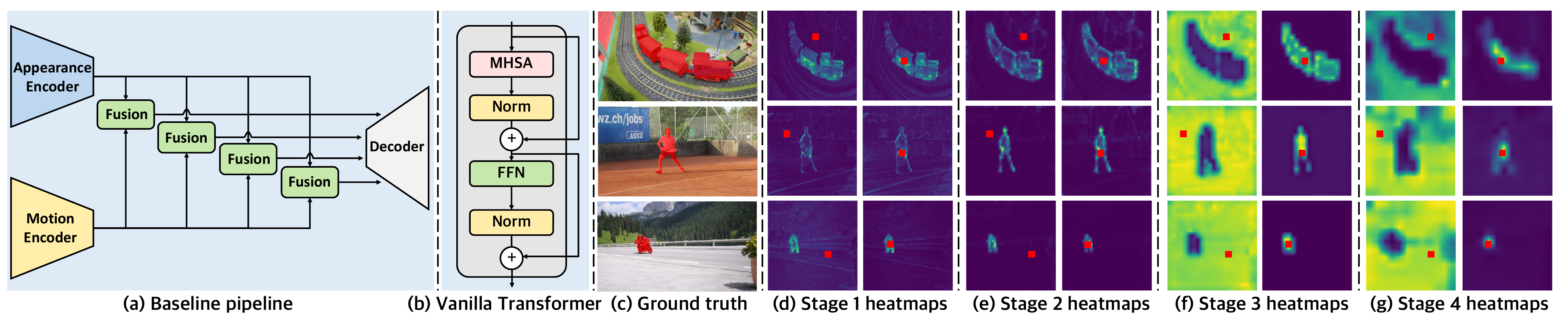}
    \captionof{figure}{Visualization of the learned attention dependencies of the baseline pipeline (a) using vanilla Transformer (b) as fusion modules. (d)-(g) present the attention maps (heatmaps) of different queries (red points) computed by the vanilla Transformer in different fusion stages: for the same input image, different queries have almost the same attention maps in low-level stages (d) (e), while they only focus on image regions with the same semantics in high-level stages (f) (g).}
\label{attn_maps}
\end{center}
}]

\maketitle
\ificcvfinal\thispagestyle{empty}\fi

\begin{abstract}
{\let\thefootnote\relax\footnotetext{\noindent * Corresponding author: wyfan@dlut.edu.cn.}}

    Recent leading zero-shot video object segmentation (ZVOS) works devote to integrating appearance and motion information by elaborately designing feature fusion modules and identically applying them in multiple feature stages. Our preliminary experiments show that with the strong long-range dependency modeling capacity of Transformer, simply concatenating the two modality features and feeding them to vanilla Transformers for feature fusion can distinctly benefit the performance but at a cost of heavy computation. Through further empirical analysis, we find that \textbf{attention dependencies learned in Transformer in different stages exhibit completely different properties}: global query-independent dependency in the low-level stages and semantic-specific dependency in the high-level stages. Motivated by the observations, we propose two Transformer variants: i) Context-Sharing Transformer (CST) that learns the global-shared contextual information within image frames with a lightweight computation. 
    ii) Semantic Gathering-Scattering Transformer (SGST) that models the semantic correlation separately for the foreground and background and reduces the computation cost with a soft token merging mechanism.
    We apply CST and SGST for low-level and high-level feature fusions, respectively, formulating a \textbf{level-isomerous Transformer} framework for ZVOS task. Compared with the baseline that uses vanilla Transformers for multi-stage fusion, ours significantly increase the speed by $\mathbf{13\times}$ and achieves new state-of-the-art ZVOS performance. 
    Code is available at \href{https://github.com/DLUT-yyc/Isomer}{https://github.com/DLUT-yyc/Isomer}.
\end{abstract}

\section{Introduction}
Zero-shot Video Object Segmentation (ZVOS) aims at discovering the most visually attractive objects in a video sequence and serves as a fundamental computer vision technique. Different from image segmentation that mainly relies on static appearance features, ZVOS further explores temporal motion information to achieve reliable and temporally consistent results.
One popular pipeline \cite{HFAN, FSNet, AMCNet, RTNet} is integrating appearance and motion information by identically applying feature fusion modules in multiple stages as shown in Fig. \ref{attn_maps} (a). While great efforts have been made, designing effective multi-stage appearance-motion fusion approaches for ZVOS is still an open problem.

Transformers \cite{Transformer} have made remarkable breakthroughs in many computer vision tasks \cite{vit, DETR, Segformer, SwinTransformer} due to its strong capability in modeling long-range dependencies and unique flexibility for cross-modal feature fusion. Nevertheless, their merits have not been fully explored in the ZVOS field. 
A straightforward way is adopting Transformer blocks as the appearance-motion fusion modules. 
In our preliminary experiments, for each feature level, we concatenate the extracted appearance and motion features and feed them to a vanilla Transformer block (Fig. \ref{attn_maps} (b)). 
It shows that such a simple baseline achieves superior performance than all prior elaborate-designed approaches but at a cost of heavy computation. These motivate us to further investigate: 1) what Transformers exactly learn for performance gain, and 2) how to further relieve the computational burden without performance loss under this baseline framework. 

\begin{figure}[tbp]
\centering
\tiny
\includegraphics[width=\linewidth]{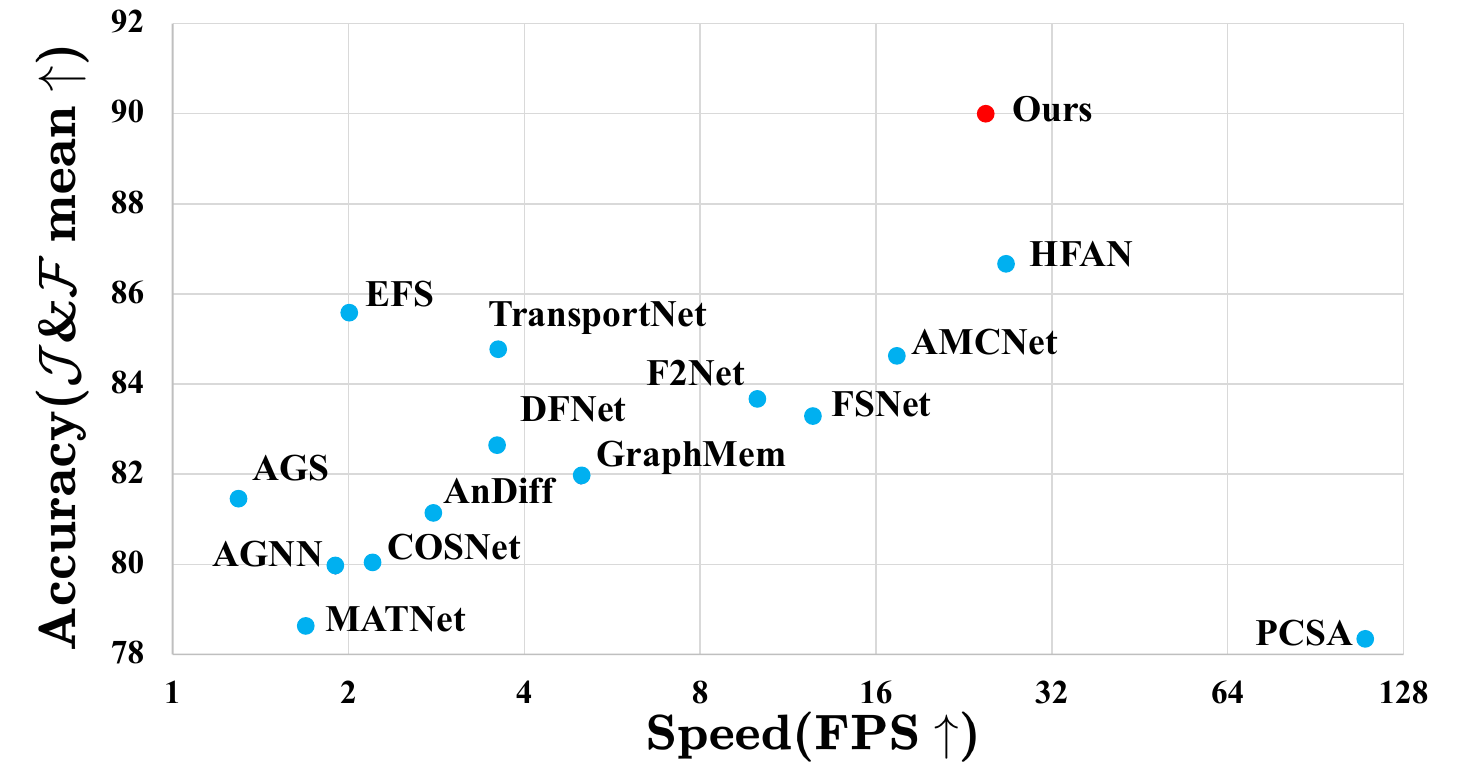}
\vspace{-2mm}
\caption{Performance \emph{v.s.} inference speed on DAVIS-16.}
\label{figure1}
\end{figure}

To answer the above questions, we visualize the attention dependencies computed by the Multi-Head Self Attention (MHSA) step of Transformers in all feature fusion stages. 
Surprisingly, it finds that \emph{vanilla Transformers in different levels characterize the attention dependencies from different perspectives to fit the ZVOS task}, which motivates our design of Transformer-based ZVOS framework as follows.  

\textbf{First, Transformers in early fusion stages only capture global query-independent dependency.} As shown in Fig.~\ref{attn_maps} (d)(e), the attention maps of different query positions are almost the same for the low-level stages, which mainly highlight the foreground object and some background contours.
It indicates that the network tends to understand the scenes via global context modeling in shallow layers and tries to distinguish the boundary between foreground and background under the ZVOS setting.
Inspired by \cite{gcnet}, we propose to simplify vanilla Transformer by computing one query-independent dependence map for all query tokens, thus modeling the global-shared contextual information and largely reducing the computational cost. We term the simplified Transformer as Context-Sharing Transformer (CST). 

\textbf{Second, Transformers in late fusion stages capture long-range semantic-specific dependency.} As shown in Fig.~\ref{attn_maps}(f)(g), the query tokens mainly pay attention to the image regions with the same semantic category, \ie foreground or background. Besides, the attention maps of the query tokens with the same semantics share many similarities, indicating existing much attention redundancy that could be further pruned. Based on these observations, we propose another Transformer variant named Semantic Gathering-Scattering Transformer (SGST), which computes foreground and background attentions separately for the corresponding query tokens with some selected representative key/value tokens. In addition, a soft token merging mechanism is adopted to enable the token selection process differentiable. Compared to the vanilla Transformer, the proposed SGST is able to model the semantic-specific dependency more explicitly and more efficiently.

Upon the above findings and the two proposed Transformer blocks, a level-\textbf{Iso}merous Transfor\textbf{mer} (Isomer) ZVOS framework is formulated by applying CST and SGST blocks for early and late fusion stages, respectively.
Compared with the baseline network that applies vanilla Transformer uniformly for all the feature fusion stages, ours treats the different fusion levels distinctively based on the observed Transformers properties, and achieves better segmentation results with $\mathbf{13\times}$ inference speed.
Compared with the existing ZVOS works, our method equipped with Swin-Tiny\cite{SwinTransformer} backbone (a comparable model size to ResNet50\cite{resnet}) obtains significantly superior performance with real-time inference (see Fig. \ref{figure1}). 

The main contributions of this work are as follows:

1) We analyze the properties of vanilla Transformers in terms of attention dependencies hierarchically learned from the ZVOS task, and propose two Transformer variants, \ie Context-Sharing Transformer (CST) and Semantic Gathering-Scattering Transformer (SGST), to model the contextual dependencies from different levels effectively and efficiently.

2) We propose a level-isomerous Transformer paradigm for the ZVOS task, which applies the developed CST and SGST for low-level early fusion and high-level late fusion, respectively. Different from the prior works that fuse appearance-motion information for all stages in an identical way, ours performs different fusion levels differentially and better fits the properties of the ZVOS network.  

3) Extensive experiments demonstrate the superiority of our method compared to the existing works as well as the strong vanilla Transformer-based baseline. To our best knowledge, this is the first successful attempt at developing a real-time Transformer-based work in the ZVOS field. 

\section{Related Work}
\subsection{Zero-shot Video Object Segmentation}
Zero-shot video object segmentation (ZVOS) aims to automatically segment the salient objects from videos without any manual prompt. It has witnessed rapid progress with the development of deep learning techniques and the establishment of large-scale datasets \cite{youtubevos, DAVIS}. Early CNN-based methods \cite{LVOS_Visual_Memory, AGS, SSAV} usually use recurrent neural networks to capture long-term dependencies. Inspired by the attention mechanism, several works \cite{COSNet, AGNN, Anchor-Diffusion} explore global context corrections between frames by designing cross-attention operations. 
Recent leading works \cite{MATNet, FSNet, AMCNet, HFAN, AMP} combine the appearance information with the motion cues extracted by the off-the-shelf optical flow methods \cite{flownet2,pwcnet,raft} and have gained significant performance improvement.
Among them, \cite{FSNet} designs a relational cross-attention module to achieve bi-directional message propagation in the appearance and motion subspaces.
HFAN \cite{HFAN} proposes a sequential feature alignment module and a feature adaptation module for appearance and motion feature alignment. 
While promising performance has been achieved, the spirit of Transformers has not been fully explored in the existing ZVOS methods. In this work, we study the properties of Transformers in ZVOS setting in-depth, and propose two novel Transformer blocks and a level-isomerous Transformer framework, hoping to provide some new insights for this field. 

\subsection{Video Salient Object Detection}
The task of Video Salient Object Detection (VSOD) is similar to ZVOS. The difference is that the ZVOS model predicts a binarized segmentation mask, while the VSOD model predicts a continuous-valued probabilistic saliency map\cite{GateNet, GateNetv2, danet, HDFNet, CAVER}.
Most works resort to capturing the temporal information by using recurrent neural networks \cite{PDB,FGRNE,SSAV} or the inter-frame motion cues \cite{MGA,DCFNet}. 
Similar to ZVOS, the prior VSOD works are also mainly based on convolutional neural networks and non-local networks, and our main contributions have not been explored in this field. 

\subsection{Lightweight Transformer in Vision Tasks}
Transformer \cite{Transformer} is first proposed for sequence-to-sequence machine translation \cite{bert, yang2019xlnet}. 
Recently, it is successfully migrated into many computer vision tasks such as image classification \cite{vit, SwinTransformer, pvt}, object detection \cite{DETR, deformable-detr}, image segmentation \cite{setr, Segformer, segmenter}, \emph{etc}.
However, the enormous computational load and memory usage make Transformer difficult to be deployed, especially for video dense prediction tasks (\emph{e.g.}, VOS and VSOD). 
To address this problem, many research efforts are taken for lightweight Transformers.
\cite{efficientformer, mobilevit, edgevit, SwinTransformer, lvt, levit, SSTVOS, STT}. Among them, one direction is to combine lightweight CNN and attention mechanism to form a hybird architecture \cite{levit, mobileformer, mobilevit, topformer}. Another track is to reduce the quadritic computation complexity of the attention mechanism \cite{litv2, lvt, pvt, pvtv2, poolformer, edgevit}. Wang \etal~\cite{pvt} design a spatial-reduction attention to reduce resource consumption. 
\cite{STT} reduces the computational complexity of Transformer by query/key selection strategy and local attention computation.
\cite{SSTVOS} proposes a sparse Transformer formulation using grid attention and strided attention for video segmentation.
As opposed to the above works that rely on hand-designed rules for Transformer lightening in a data-agnostic manner, we simplify vanilla Transformer based on the observed properties of Transformers under the ZVOS setting, which is able to learn task-aware attention in a more flexible data-driven way, yielding both accuracy and efficiency gain for ZVOS problem (see Sec. \ref{sec:ablation}).

The most relevant work for ours is GCNet \cite{gcnet}, which proposes a global context block to simplify the non-local network \cite{non-local-network}. However, our work has significant differences compared to GCNet. First, GCNet only visualizes the high-level non-local neural networks and proposes to learn the query-independent attention map for all query positions. On the contrary, we analyze both low-level and high-level Transformers under the ZVOS task, and find that the attention dependencies modeled by Transformer are totally different along the network depth: global query-independent dependency in the low-level stages and semantic-specific dependency in the high-level stages. Second, it should be recognized that our CST is inspired by GCNet. Nevertheless, the design of SGST and our main insight about the level-isomerous framework are unique.

\begin{figure*}[t]
\centering
\footnotesize
\includegraphics[width=\linewidth]{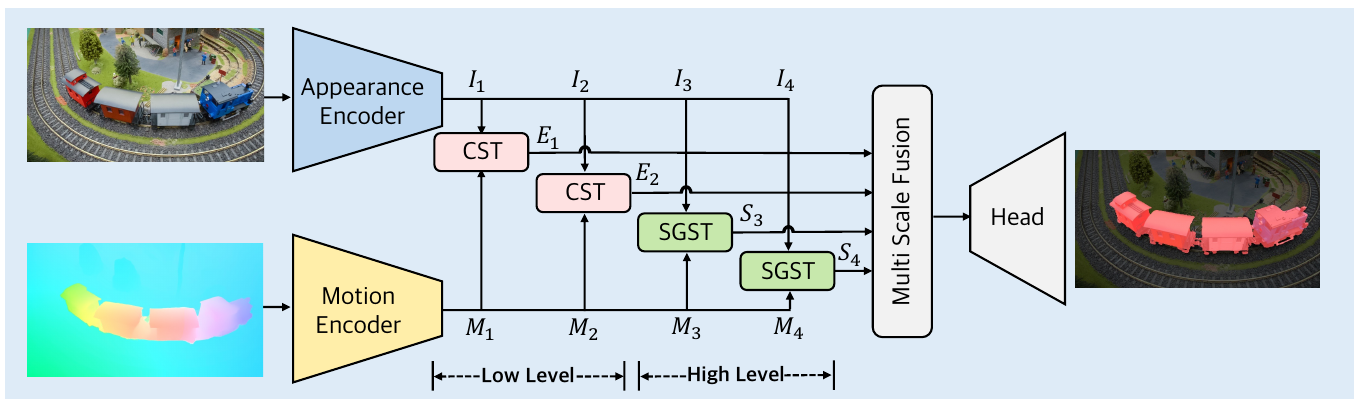}
\caption{Overview of the proposed framework (Isomer). Given the current video frame and its corresponding optical flow map, Isomer extracts hierarchical appearance and motion features by two backbones. Then the proposed \textbf{CST (Context-Sharing Transformer)} and \textbf{SGST (Semantic Gathing-Scattering Transformer)} are adopted for cross-modality feature fusion in the low-level (the first two) and high-level (the last two) stages, respectively. The multi-stage fused features are fed into a segmentation head to obtain the final result.} 
\label{pipeline}
\end{figure*}

\section{Methodology}
\subsection{Vanilla Transformer Baseline}
The baseline method is designed following the commonly adopted framework \cite{FSNet,HFAN} as shown in Fig. \ref{attn_maps} (a). It consists of an appearance backbone, a motion backbone, multiple fusion modules, and a decoder. 

Given the current video frame and the optical flow map that is computed between the current frame with its adjacent one, 
the appearance backbone and motion backbone extract four-stage \emph{appearance features} $\mathbf{I}_l$ and \emph{motion features} $\mathbf{M}_l$ ($l\in\{1,2,3,4\}$), respectively.
For each stage, one fusion module is applied to integrate appearance and motion features. 
Specifically, the extracted two modality features ($\mathbf{I}_l$ and $\mathbf{M}_l$) are first channel-wise combined to obtain a mixing representation $\mathbf{X}_l \in \mathbb{R}^{C \times H \times W}$, where $C, H, W$ denote the channel number, height, and width of $\mathbf{X}_l$, respectively. It can also be regarded as a list of tokens $\{\mathbf{x}_{i} | \mathbf{x}_{i} \in \mathbb{R}^{C}, i=1,2,\ldots, N\}$ with $N = H \times W$. Then, the mixing representation is fed into a vanilla Transformer block (see Fig. \ref{attn_maps}(b)) for cross-modality feature fusion.
Following previous works \cite{HFAN,AMCNet,FSNet}, we use a feature pyramid decoder \cite{Segformer} to leverage the four-stage fused features for the final segmentation prediction.

Without bells and whistles, the vanilla Transformor-based baseline achieves outstanding performance but also brings about a heavy computational burden, which motivates us to further explore an effective solution to making a trade-off between performance and computation.

\begin{figure}[t]
\centering
\small
\includegraphics[width=\linewidth]{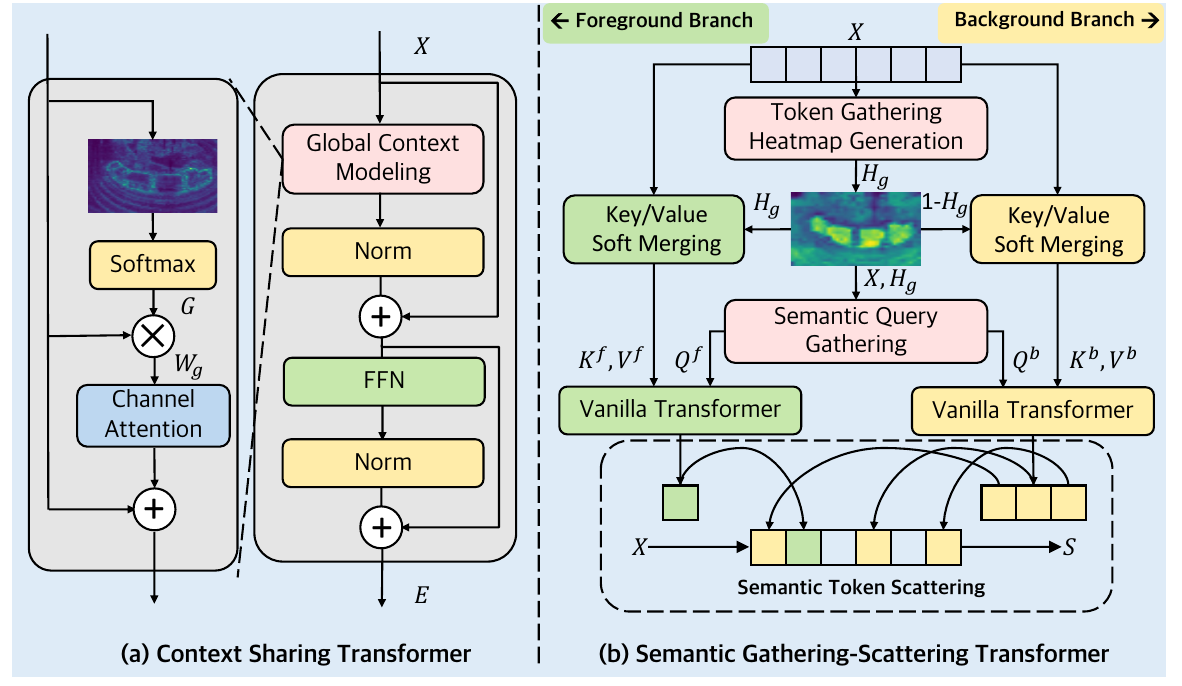}
\caption{Illustration of the proposed CST and SGST modules.}
\label{module}
\end{figure}

\subsection{Isomerous Transformer for ZVOS}

Driven by the observations in Fig. \ref{attn_maps}, we reform the vanilla Transformer fusion blocks in the baseline by developing two new Transformer blocks, named Context-Sharing Transformer (CST) and Semantic Gathering-Scattering Transformer (SGST). We apply them for multi-level feature fusion following the observed network properties and thus form a new level-isomerous Transformer (Isomer) framework. In the following, we will first illustrate the proposed blocks, followed by an overview of the framework.

\subsubsection{Context Sharing Transformer}
The Context-Sharing Transformer (CST) is developed to simplify the MHSA step in vanilla Transformer with a global context modeling as shown in Fig. \ref{module} (a), which computes global query-independent attention for all queries. 

The global context modeling step first performs a query-shared spatial-wise attention followed by a channel-wise attention. Specifically, given a mixing representation $\mathbf{X}_l \in \mathbb{R}^{C \times H \times W}$ from one low-level stage $l (l\in\{1,2\})$, it first generates a one-channel attention weight map $\mathbf{G}_l \in \mathbb{R}^{H \times W}$ by a $1 \times 1$ convolution with a Softmax function, which is used to weight $\mathbf{X}_l$ to obtain a query-shared weighted representation $\mathbf{W}_l^g \in \mathbb{R}^{C}$.
Then two $1 \times 1$ convolution layers interleaved by BN and ReLU are adopted to refine the global context features, which can be seen as a type of channel attention. A skip connection is adopted at the end of the global context modeling step to aggregate the global context information with $\mathbf{X}_l$. 
The aggregated result is sent to the remaining components of vanilla Transformer and produces the final fused feature $\mathbf{E}_l \in \mathbb{R}^{C \times H \times W}$.

While being embarrassingly simple, CST significantly speeds up the baseline inference (36 FPS \emph{v.s.} 3 FPS) with negligible performance drop when replacing vanilla Transformers with CST for the first two fusion stages. 

\subsubsection{Semantic Gathering-Scattering Transformer}
\label{SGST}
Semantic Gathering-Scattering Transformer (SGST) is to explicitly model the foreground/background semantic dependencies while reducing the attention redundancy.  
As shown in Fig. \ref{module} (b), SGST consists of two parallel branches to separately process foreground and background, mainly including semantic query gathering, key/value soft merging, dependencies calculation with a standard Transformer step, and semantic token scattering. As the two branches share a similar process, we simplify the description by illustrating only the foreground branch. 

\noindent \textbf{Semantic Query Gathering.} Given a mixing representation $\mathbf{X}_l \in \mathbb{R}^{C \times H \times W}$ from one high-level stage $l (l\in\{3,4\})$, a one-channel token gathering heatmap $\mathbf{H}_l^g \in \mathbb{R}^{H \times W}$ is firstly generated using a $1\times 1$ convolution layer followed by a Sigmoid function. Upon the heatmap, the foreground tokens are identified via $\mathbf{X}_l[h_{i} >= 0.5]$, and gathered to form a list of foreground queries $\mathbf{Q}_l^{f}$. $h_{i}$ denotes the heatmap value at position $i$, and 0.5 is the threshold to distinguish foreground from background\footnote{The background queries $\mathbf{Q}_l^{b}$ are obtained using $\mathbf{X}_l[h_{i} < 0.5]$.}.

\noindent \textbf{Key/Value Soft Merging.} We begin by calculating the dot product between $\mathbf{X}_l$ and $\mathbf{H}_l^g$\footnote{The background branch adoptes $1-\mathbf{H}_l^g$.}, producing a heatmap enhanced foreground token sequence $\mathbf{X}_l^{e} \in \mathbb{R}^{C \times N}$, where $N=H \times W$. To adaptively mine the representative information and remove redundancy, we merge the $N$ tokens of $\mathbf{X}_l^{e}$ into $K$ tokens $\mathbf{X}_l^{c} \in \mathbb{R}^{C \times K}$ ($K << N$) in a soft way, which is achieved using a learnable transformation matrix $\mathbf{W}_l^m \in \mathbb{R}^{N \times K}$ applied on $\mathbf{X}_l^{e}$. Then the compact foreground key and value sequences $\mathbf{K}_l^{f}, \mathbf{V}_l^{f} \in \mathbb{R}^{C\times K}$ are generated with linear transformations of $\mathbf{X}_l^{c}$.   

\noindent \textbf{Dependencies Calculation.} With the compressed foreground query $\mathbf{Q}_l^{f}$, key $\mathbf{K}_l^{f}$, and value $\mathbf{K}_l^{f}$, a standard Transformer block is adopted to model the semantic dependencies and update the corresponding foreground representation tokens, where the attention computation can be largely reduced at the same time.

\noindent \textbf{Semantic Token Scattering.} With the original mixing representation $\mathbf{X}_l$, the updated foreground and background tokens are scattered back according to the indexes in the query gathering step and obtain the final fused feature $\mathbf{S}_l$.

In our experiments, $K$ is set to $\frac{1}{9}N$ in the key/value soft merging step, reducing $87\%$ of computation of MHSA in total compared to vanilla Transformer.
SGST's unique ability of explicitly modeling semantic dependencies through foreground-background query separation, along with its efficient token merging mechanism, allows it to dramatically reduce computational complexity while maintaining top-notch performance.

\subsubsection{Isomerous Framework}
With the two proposed Transformer variants, we take the vanilla Transformer baseline one step further. As shown in Fig. \ref{pipeline}, we replace vanilla Transformers in the first two stages with our Context Sharing Transformer (CST) to model the global-shared contextual information within image frames, while in the last two stages with our Semantic Gathering-Scattering Transformer (SGST) to models the semantic correlation explicitly. 
Consequently, it formulates a level-isomerous Transformer (Isomer), which treats the different fusion levels distinctively based on the observed properties of Transformers to better fit the ZVOS task. 

\begin{table*}[htbp]
  \centering
  \caption{ZVOS performance on DAVIS-16 validation set. ``CRF'' means that conditional random field \cite{crf} is applied as post-processing. The inference speed is tested on one 3090 GPU. The best and second-best scores are indicated in \textcolor[rgb]{1,0,0}{red} and \textcolor[rgb]{0,0,1}{blue}, respectively.}
    \begin{tabular}{c|c|c|c|c|c|c|c}
    \hline
    Method & Publication & Backbone  & CRF   & $\mathcal{J}$ Mean $\uparrow$ & $\mathcal{F}$ Mean $\uparrow$ & $\mathcal{J}$\&$\mathcal{F}$ $\uparrow$ & FPS $\uparrow$ \\
    \hline
    PDB\cite{PDB}   & ECCV2018 & ResNet-50   & $\checkmark$     & 77.2  & 74.5  & 75.9 &20.0\\
    AGS\cite{AGS}   & CVPR2019 & ResNet-101    & $\checkmark$     & 79.7  & 77.4  & 78.6 &1.7\\
    AGNN\cite{AGNN}  & ICCV2019 & ResNet-101    & $\checkmark$     & 80.7  & 79.1  & 79.9 &1.9\\
    COSNet\cite{COSNet} & CVPR2019 & ResNet-101   & $\checkmark$     & 80.5  & 79.5  & 80.0 &2.2\\
    AnDiff\cite{Anchor-Diffusion} & CVPR2019 & ResNet-101      &       & 81.7  & 80.5  & 81.1 &2.8\\
    MATNet\cite{MATNet} & AAAI2020 & ResNet-101    & $\checkmark$     & 82.4  & 80.7  & 81.5 &1.3\\
    GraphMem\cite{graphmem}  & ECCV2020 & ResNet-50    & $\checkmark$     & 82.5  & 81.2  & 81.9 &5.0\\
    FSNet\cite{FSNet} & ICCV2021 & ResNet-50    & $\checkmark$     & 83.4  & 83.1  & 83.3 &12.5 \\
    AMCNet\cite{AMCNet} & ICCV2021 & ResNet-101 & $\checkmark$     & 84.5  & 84.6  & 84.6 & 17.5\\
    RTNet\cite{RTNet} & CVPR2021 & ResNet-101   & $\checkmark$     & 85.6  & 84.7  & 85.2 &-\\
    HFAN\cite{HFAN}  & ECCV2022 & Swin-Tiny    &       & \textcolor[rgb]{ 0,0,1}{86.0} & \textcolor[rgb]{ 0,0,1}{87.3} & \textcolor[rgb]{ 0,0,1}{86.7} & \textcolor[rgb]{ 1,0,0}{26.7} \\
    Ours  &   -   & Swin-Tiny  &       & \textcolor[rgb]{ 1,  0,  0}{88.8} & \textcolor[rgb]{ 1,  0,  0}{91.1} & \textcolor[rgb]{ 1,  0,  0}{90.0} & \textcolor[rgb]{ 0,0,1}{24.6}\\
    \hline
    \end{tabular}%
  \label{davis_performance}%
\end{table*}%

\begin{table*}[htbp]
  \centering
  \caption{ZVOS performance on FBMS validation set.}
    \begin{tabular}{c|cccccccccc}
    \hline
    \thead{Method} 
    & \thead{OBN \\ \cite{OBN}}
    & \thead{PDB \\ \cite{PDB}}
    & \thead{COSNet \\ \cite{COSNet}}
    & \thead{MATNet \\ \cite{MATNet}}
    & \thead{AMCNet \\ \cite{AMCNet}}
    & \thead{APS \\ \cite{APS}}
    & \thead{F2Net \\ \cite{F2Net}}
    & \thead{EFS \\ \cite{EFS}}
    & \thead{TransportNet \\ \cite{TransportNet}}
    & \thead{Ours} \\
    \hline
    $\mathcal{J}$ Mean $\uparrow$ & 73.9  & 74    & 75.6  & 76.1  & 76.5 & 76.7  & 77.5  & 77.5  & 78.7  & \textcolor[rgb]{ 1,  0,  0}{87.6} \\
    \hline
    \end{tabular}%
  \label{fbms_performance}%
\end{table*}%

\begin{table}[htbp]
  \centering
  \small
  \caption{ZVOS performance on Long Videos dataset.}
    \begin{tabular}{c|cc|c}
    \hline
    Method & $\mathcal{J}$ Mean $\uparrow$ & $\mathcal{F}$ Mean $\uparrow$ & $\mathcal{J}$\&$\mathcal{F}$ $\uparrow$ \\
    \hline
    3DCSeg\cite{3DCSeg} & 34.2  & 33.1  & 33.7 \\
    MATNet\cite{MATNet} & 66.4  & 69.3  & 67.9 \\
    AGNN\cite{AGNN} & 68.3  & 68.6  & 68.5 \\
    HFAN\cite{HFAN} & 80.2  & 83.2  & 81.7 \\
    Ours  & \textcolor[rgb]{1,0,0}{81.4}  & \textcolor[rgb]{1,0,0}{84.9}  & \textcolor[rgb]{1,0,0}{83.2} \\
    \hline
    \end{tabular}%
  \label{long_videos}%
\end{table}%

\begin{table*}[htbp]
  \centering
  \small
  \caption{Overall VSOD performance on three benchmark datasets. The best and second-best scores are indicated in \textcolor[rgb]{1,0,0}{red} and \textcolor[rgb]{0,0,1}{blue}.}
    \begin{tabular}{c|cccc|cccc|cccc}
    \hline
    \multirow{2}[4]{*}{Methods} & \multicolumn{4}{c|}{DAVIS-16\cite{DAVIS}}    & \multicolumn{4}{c|}{FBMS\cite{FBMS}}     & \multicolumn{4}{c}{MCL\cite{MCL}} \\
          &  $\mathcal{M}$↓    &  $\mathcal{E}_\xi^{max}$↑    &  $\mathcal{F}_\beta^{max}$↑    & $\mathcal{S}_\alpha$↑    
          &  $\mathcal{M}$↓    &  $\mathcal{E}_\xi^{max}$↑    &  $\mathcal{F}_\beta^{max}$↑    & $\mathcal{S}_\alpha$↑    
          &  $\mathcal{M}$↓    &  $\mathcal{E}_\xi^{max}$↑    &  $\mathcal{F}_\beta^{max}$↑    & $\mathcal{S}_\alpha$↑ \\
    \hline
    RCR\cite{RCR} & 0.027  & 0.947  & 0.848  & 0.886  & 0.053  & 0.905  & 0.859  & 0.872  & 0.028  & 0.895  & 0.742  & \textcolor[rgb]{0, 0, 1}{0.864 } \\
    SSAV\cite{SSAV}  & 0.028  & 0.948  & 0.861  & 0.893  & 0.040  & 0.926  & 0.865  & 0.879  & 0.026  & 0.889  & 0.773  & 0.819  \\
    MGA\cite{MGA}   & 0.022  & 0.961  & \textcolor[rgb]{0, 0, 1}{0.902}  & 0.913  & \textcolor[rgb]{0, 0, 1}{0.027 } & \textcolor[rgb]{0, 0, 1}{0.949 } & \textcolor[rgb]{0, 0, 1}{0.910 } & \textcolor[rgb]{0, 0, 1}{0.907 } & 0.031  & 0.901  & 0.798  & 0.845  \\
    PCSA\cite{PCSA}  & 0.022  & 0.961  & 0.880  & 0.902  & 0.041  & 0.914  & 0.831  & 0.866  & N/A   & N/A   & N/A   & N/A \\
    DCFNet\cite{DCFNet} & \textcolor[rgb]{0, 0, 1}{0.016 } & 0.969  & 0.900  & 0.914  & 0.037  & 0.916  & 0.849  & 0.877  & 0.029  & 0.875  & 0.716  & 0.762  \\
    FSNet\cite{FSNet} & 0.020  & \textcolor[rgb]{0, 0, 1}{0.970 } & \textcolor[rgb]{0, 0, 1}{0.902 } & \textcolor[rgb]{0, 0, 1}{0.920 } & 0.041  & 0.935  & 0.888  & 0.890  & \textcolor[rgb]{0, 0, 1}{0.023 } & \textcolor[rgb]{0, 0, 1}{0.924 } & \textcolor[rgb]{0, 0, 1}{0.821 } & \textcolor[rgb]{0, 0, 1}{0.864 } \\
    Ours  & \textcolor[rgb]{ 1,  0,  0}{0.010 } & \textcolor[rgb]{ 1,  0,  0}{0.987 } & \textcolor[rgb]{ 1,  0,  0}{0.946 } & \textcolor[rgb]{ 1,  0,  0}{0.950 } & \textcolor[rgb]{ 1,  0,  0}{0.019 } & \textcolor[rgb]{ 1,  0,  0}{0.974 } & \textcolor[rgb]{ 1,  0,  0}{0.944 } & \textcolor[rgb]{ 1,  0,  0}{0.934 } & \textcolor[rgb]{ 1,  0,  0}{0.015 } & \textcolor[rgb]{ 1,  0,  0}{0.967 } & \textcolor[rgb]{ 1,  0,  0}{0.882 } & \textcolor[rgb]{ 1,  0,  0}{0.893 } \\

    \hline
    \end{tabular}%
  \label{vsod}%
\end{table*}%

\subsection{Implement Details} \label{sec:implement}

We use Swin-Tiny \cite{SwinTransformer} as our backbone when reporting the final results for fair comparison, which has a comparable model size with ResNet50\cite{resnet}. Other backbones are also compared in our experiments.
Following \cite{HFAN,AMCNet}, the well-trained RAFT \cite{raft} is adopted to generate the optical flow maps for the video data. All the input images are resized to a spatial resolution of $512 \times 512$. Data augmentation including horizontal flipping and photometric distortion is adopted during training. 
We utilize a subset of the Youtube-VOS \cite{youtubevos} training set (1 frame per every 30 frames sampled) to pre-train the network based on \cite{MATNet, RVOS, 3DCSeg}, followed by network fine-tuning with DAVIS-16 \cite{DAVIS} and FBMS \cite{FBMS} training sets.
The AdamW \cite{AdamW} optimizer is adopted with a fixed learning rate of 6e-5 throughout the training process. 
Our network is end-to-end trained on one NVIDIA 3090 GPU with a mini-batch size of 8, using binary cross-entropy loss for supervision.

\section{EXPERIMENT}
\subsection{Datasets and Evaluation Metrics}

\noindent \textbf{Datasets.} We perform evaluation on four widely adopted ZVOS datasets: DAVIS-16 \cite{DAVIS}, FBMS \cite{FBMS}, Long-Videos \cite{AFB-URR} and Youtube Objects\cite{youtubeobjects}. 
In addition, we also conduct experiments for the video salient object detection (VSOD) task to comprehensively evaluate our method using three datasets, including DAVIS-16, FBMS, and MCL \cite{MCL}. 

\noindent \textbf{Evaluation metrics.} We report the standard evaluation metrics for ZVOS task, including mean of region similarity ($\mathcal{J}$ Mean), mean of contour accuracy ($\mathcal{F}$ Mean) \cite{DAVIS}, and $\mathcal{J} \& \mathcal{F}$ that is computed by averaging $\mathcal{J}$ Mean and $\mathcal{F}$ Mean.
For VSOD task, we adopt four widely used metrics: MAE \cite{MAE-metric} ($\mathcal{M}$), maximum E-measure ($\mathcal{E}_\xi^{max}$) \cite{Emeasure}, maximum F-measure ($\mathcal{F}_\beta^{max}$, $\beta^{2}=0.3$) \cite{Fmeasure}, and S-measure ($\mathcal{S}_\alpha$, $\alpha=0.5$) \cite{Smeasure}.  
Please note that the predicted maps are binarized with a threshold of 0.5 for ZVOS evaluation, but not for VSOD evaluation following \cite{FSNet, DCFNet, MGA}. 

\subsection{Comparison with State-of-the-art}

\noindent \textbf{Evaluation on ZVOS.} Tab. \ref{davis_performance}, \ref{fbms_performance}, and \ref{long_videos} show the overall ZVOS performance on DAVIS-16, FBMS, Long-Videos dataset, respectively. 
We also report the inference speed of all the compared methods tested using a 3090 GPU. 
The proposed method Isomer consistently outperforms the compared methods on all datasets with nearly real-time inference (24.6 FPS). 
Specifically, on DAVIS-16 dataset, compared with the current leading method HFAN \cite{HFAN}, Isomer achieves a significant improvement of 2.9\% in terms of $\mathcal{J}$\&$\mathcal{F}$ with a comparable inference speed. Besides, Isomer reaches an improvement of 9.7\% in terms of $\mathcal{J}$ compared with TransportNet \cite{TransportNet} on FBMS dataset.
We also conduct evaluation on the Youtube-Objects dataset{\cite{youtubeobjects}} and our method with a $\mathcal{J}$ score of 74.6 surpasses the previous leading method HFAN (73.4 $\mathcal{J}$).
 
\noindent \textbf{Evaluation on VSOD.} Tab. \ref{vsod} provides the quantitative comparison for VSOD task, showing that Isomer achieves the best results across all evaluation metrics on the three datasets.
Compared with the second best methods, Isomer outperforms FSNet \cite{FSNet} by 5.3\% and 3.2\% in terms of $\mathcal{F}_\beta^{max}$ and $\mathcal{S}_\alpha$ on DAVIS-16 dataset, and exceeds MGA \cite{MGA} by 2.9\% and 3.7\% on FBMS dataset. 
While MCL dataset has blurry boundaries in the low-resolution frames, Isomer surpasses FSNet \cite{FSNet} by 4.7\% and 7.4\% for $\mathcal{E}_\xi^{max}$ and $\mathcal{F}_\beta^{max}$, respectively.
These results indicate that our method generalizes well across both ZVOS and VSOD tasks.

\subsection{Ablation Study} \label{sec:ablation}
To analyze the impact of our key components, we conduct several ablation studies on the DAVIS-16 validation set with ZVOS evaluation metrics. To conserve computing resources, we use a lightweight backbone MiT-b0 \cite{Segformer} for the following experiments unless otherwise stated.

\noindent \textbf{Effectiveness of CST, SGST, and Level-Isomerous Scheme.} 
We begin by studying the importance of leveraging Transformers for effective feature fusion. 
Tab. \ref{table:fusion_modules} shows that a basic vanilla Transformer (VT) implementation can surpass leading CNN-based fusion modules \cite{MGA, MATNet, HFAN} with the same backbone (MiT-b0) and experimental setup. However, the VT baseline incurs substantial computation and can only achieve a speed of 3 FPS with 4.5$\times$ FLOPs of HFAN.
Then, we replace the VT blocks with our CST in the low-level (\ie the first two) feature fusion stages, and observe a significant speed improvement (36 FPS) with almost no performance drop. Based on this, we then replace VT with our SGST in the high-level (\ie the last two) feature fusion stages and obtain further improvements in both speed and performance. 
For the amount of calculation, our Isomer greatly reduces the compute of VT-baseline from 18.2G to 4.1 GFLOPs. Compared to the current leading method HFAN (81.2 $\mathcal{J}$\&$\mathcal{F}$, 4.0 GFLOPs), ours delivers significant performance gain (85.2 $\mathcal{J}$\&$\mathcal{F}$) with comparable total compute (4.1 GFLOPs).
To further investigate the superiority of the proposed level-isomerous framework, we also apply CST or SGST identically for all the stages. It shows that the proposed level-isomerous Transformer framework can achieve the best trade-off between speed and accuracy. 

\begin{table}[t]
  \centering
  \small
  \caption{Ablation study for the proposed CST and SGST blocks, and our level-isomerous fusion scheme. \textbf{Bold} font indicates the best trade off between accuracy and speed.}
  \vspace{1mm}
    \begin{tabular}{c|c|c|c|c}
    \hline
    Low Level & High Level & $\mathcal{J}$ Mean $\uparrow$ & $\mathcal{F}$ Mean $\uparrow$ & FPS $\uparrow$\\
    \hline
    MGA\cite{MGA} & MGA\cite{MGA} & 79.7 & 80.7 & 27 \\
    MAT\cite{MATNet} & MAT\cite{MATNet} & 80.0 & 80.9 & 16 \\
    HFAN\cite{HFAN} & HFAN\cite{HFAN} & 81.5 & 80.8 & 42 \\
    VT & VT & 84.2 & 85.4 & 3 \\
    CST & VT & 84.2 & 85.2 & 36 \\
    \textbf{CST} & \textbf{SGST} & \textbf{84.6} & \textbf{85.6} & \textbf{39} \\ 
    CST & CST & 82.9 & 83.6 & 44 \\
    SGST & SGST & 84.5 & 85.8 & 29 \\
    VT & SGST & 84.6 & 85.7 & 4 \\
    SGST & CST & 82.9 & 83.9 & 33 \\
    \hline
    \end{tabular}%
  \label{table:fusion_modules}%
\end{table}%

\begin{table}[t]
  \centering
  \small
  \caption{Performance comparison with lightweight Transformers.}
    \begin{tabular}{c|c|c|c}
    \hline
    Method & $\mathcal{J}$ Mean $\uparrow$ & $\mathcal{F}$ Mean $\uparrow$ & FPS $\uparrow$ \\
    \hline
    AxialNet\cite{axialnet} & 83.6 & 84.4 & 32 \\
    PoolFormer\cite{poolformer} & 82.0 & 83.8  & 41 \\
    EdgeViT\cite{edgevit} & 84.1 & 83.9 & 36 \\
    PVT V2\cite{pvtv2} & 83.0 & 84.6 & 30 \\
    PVT\cite{pvt} & 81.0 & 83.6 & 28 \\
    \textbf{Ours} & \textbf{84.6} & \textbf{85.6}&\textbf{39} \\
    \hline
    \end{tabular}%
  \label{lightweight_transformer_compare}%
\end{table}%

\noindent \textbf{Comparison with Existing Lightweight Transformers.}
We also conduct experiments using current general lightweight Transformers as the fusion modules to further verify the superiority of our CST and SGST blocks for ZVOS task. Results are reported in Tab. \ref{lightweight_transformer_compare}. 
Compared with the VT block (fourth row in Tab. \ref{table:fusion_modules}), these lightweight Transformers all achieve notable acceleration while sacrificing accuracy. In contrast, Our method is designed based on Transformers' behavior under ZVOS settings to learn task-specific attention in a flexible data-driven manner, and provides both improved accuracy and efficiency.

\begin{table}[t]
  \centering
  \small
  \caption{Performance comparison with the state-of-the-art method HFAN\cite{HFAN} on different backbones.} 
    \begin{tabular}{c|c|c}
    \hline
    Backbone & $\mathcal{J}$\&$\mathcal{F}$ $\uparrow$ (HFAN\cite{HFAN}) & $\mathcal{J}$\&$\mathcal{F}$ $\uparrow$ (Ours) \\
    \hline
    MiT-b0 & 81.2  & \textbf{85.1} \\
    ResNet101 & 87.0    & \textbf{87.5} \\
    Swin-Tiny & 86.7  & \textbf{89.2} \\
    \hline
    \end{tabular}%
  \label{backbone_compare}%
\end{table}%

\noindent \textbf{Performance with Different Backbones.} 
We adopt different backbones to verify the generality of the proposed feature fusion method. Tab. \ref{backbone_compare} shows that our method consistency outperforms the recent leading method HFAN \cite{HFAN}, and the improvement is particularly prominent for the lightweight backbones MiT-b0 \cite{Segformer} and Swin-Tiny \cite{SwinTransformer}. 
We also conduct experiments with ResNet50 and ours achieve 85.1 $\mathcal{J}$\&$\mathcal{F}$ with 26 FPS, which is comparable in accuracy but significantly more efficient than the previous leading method {EFS\cite{EFS}} (85.6 $\mathcal{J}$\&$\mathcal{F}$, 2 FPS).

\begin{table}[t]
\centering
\small
\caption{Effectiveness of foreground-background separate modeling in SGST. ``S" denotes foreground and background separation. ``F" denotes foreground.}
\begin{tabular}{c|c|c} \hline
Method  & $\mathcal{J}$ Mean $\uparrow$ & $\mathcal{F}$ Mean $\uparrow$ \\\hline
    Ours (w/o S)  & 84.3 & 85.4 \\
    only F & 81.8 & 82.2 \\
    \textbf{Ours}    & \textbf{84.6} & \textbf{85.6} \\
    \hline
    \end{tabular}%
  \label{design_sgst}%
\end{table}%

\noindent \textbf{Foreground-background Separate Modeling in SGST.} 
In Tab. \ref{design_sgst}, we ablate our method by removing the foreground-background separation in SGST (top row), which calculates attention by indiscriminately mixing the foreground and background together as vanilla Transformer. Results verify the effectiveness of the main concept of SGST that modeling the semantic dependencies explicitly for the foreground and background.   
Furthermore, we explore the impact of modeling only the foreground dependencies, which aligns with our final goal. However, as evidenced by the second row of Tab. \ref{design_sgst}, a significant decrease in performance is observed, suggesting that both foreground and background are crucial in aiding the comprehension of semantics. 

\noindent \textbf{Merging Ratio in Soft Token Selection.}
SGST uses token soft merging to eliminate foreground/background redundancy. 
We study the impact of different merging ratios (\ie $K/N$) on performance. 
Tab. \ref{Soft ratio} shows that $1/9$ merging ratio obtains superior performance with $87\%$  reduction of computational cost compared with vanilla Transformer.

\begin{table}[t]
  \centering
  \small
  \caption{Ablations on the ratio of key/value soft merging in SGST.}
    \begin{tabular}{c|c|c}
    \hline
    $K/N$ & $\mathcal{J}$ Mean $\uparrow$ & $\mathcal{F}$ Mean $\uparrow$\\
    \hline
    1     & 83.6 & 84.1 \\
    4/9   & 84.1 & 85.2 \\
    1/4   & 84.1 & 85.4 \\
    \textbf{1/9}   & \textbf{84.6} & \textbf{85.6} \\
    1/36  & 83.7 & 84.4 \\
    \hline
    \end{tabular}%
  \label{Soft ratio}%
\end{table}%

\begin{figure*}[t]
\centering
\footnotesize
\includegraphics[width=\linewidth]{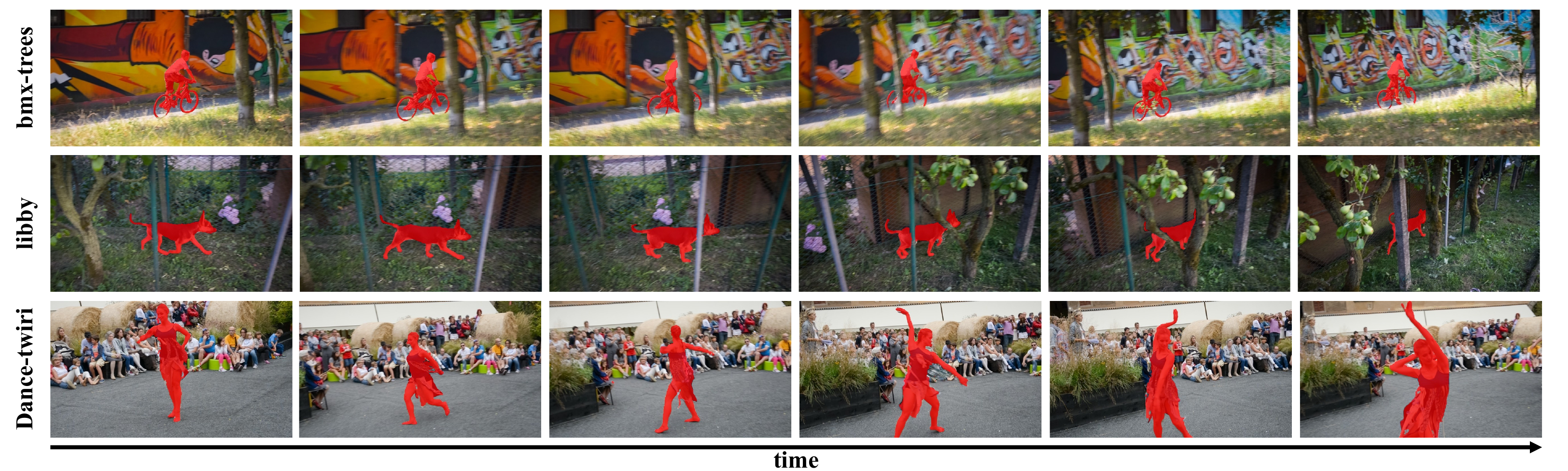}
\caption{Qualitative results on three challenging video clips from DAVIS-16\cite{DAVIS}.}
\label{results_visualization}
\end{figure*}

\begin{figure}[t]
\centering
\small
\includegraphics[width=\linewidth]{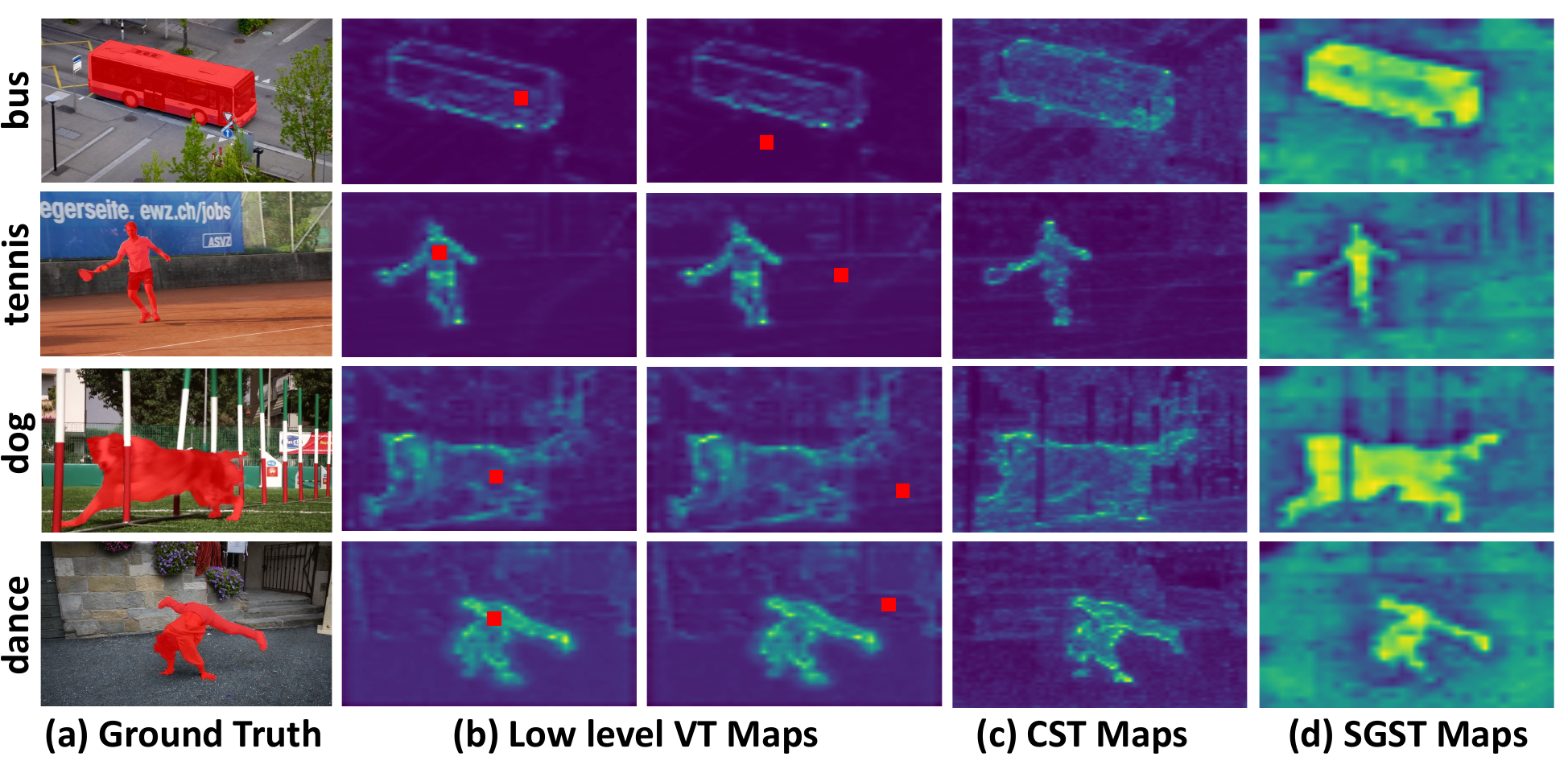}
\caption{Illustration of the attention maps from VT and CST and token gathering heatmaps from SGST.}
\label{heatmap}
\end{figure}

\noindent \textbf{Visualization.} 
Fig. \ref{results_visualization} shows some qualitative results of Isomer, showing its promising prediction ability across various challenging situations. 
Fig. \ref{heatmap} visualizes several attention maps (heatmaps) of VT (b) and CST (c), showing that CST captures mostly similar query-independent dependency as VT but with less computational effort.
Fig. \ref{heatmap}(d) shows the token gathering heatmaps learned in SGST, which
effectively distinguishes between foreground and background to help the network explicitly
model semantic dependencies.

\noindent \textbf{Limitation and Future Work.} Since videos often contain multiple moving objects, optical flow maps tend to have a lot of noise. Therefore, directly fusing optical flow information at the pixel level may not be the best approach for ZVOS. Instead, a promising solution could be a simple coarse-to-fine pipeline, which involves using a detection network to roughly locate the target area (such as a bounding box) through the optical flow map, and then using an image segmentation network to segment the foreground and background in the detected region. This pipeline not only helps locate moving objects with the aid of optical flow maps, but also avoids the influence of optical flow noise during segmentation, and benefits from more image segmentation data (e.g., salient object detection datasets). We plan to explore this avenue in our future work.

\section{Conclusion}
This paper proposes a new ZVOS framework named Isomer (level-\textbf{iso}merous Transfor\textbf{mer}), which treats the different feature fusion levels distinctively. We develop two core components Context-Sharing Transformer (CST) and Semantic Gathering-Scattering Transformer (SGST) to model the contextual dependencies from different levels effectively and efficiently. 
In early fusion stages, CST models the global-shared contextual information by computing one query-independent dependence map for all query tokens. In late fusion stages, SGST captures long-range semantic-specific dependency by computing the foreground and background attentions separately with a soft token merging mechanism. 
Experimental results show that the proposed Isomer achieves state-of-the-art performance in both ZVOS and VSOD tasks with real-time inference. To our best knowledge, this work is the first successful application of Transformers in ZVOS task, which could provide a new paradigm for the dense prediction vision tasks in exploring Transformer based architecture.

\section*{Acknowledgement}

The paper is supported in part by the National Key R\&D Program of China (2018AAA0102001), National Natural Science Foundation of China (62006036, 62276045, 62293542, U1903215), Dalian Science and Technology Talent Innovation Support Plan (2022RY17), and Fundamental Research Funds for Central Universities (DUT22LAB124, DUT22ZD210, DUT21RC(3)025, DUT22QN228).

{\small
\bibliographystyle{ieee_fullname}
\bibliography{egbib}

\begin{thebibliography}{10}\itemsep=-1pt

\bibitem{Fmeasure}
Radhakrishna Achanta, Sheila Hemami, Francisco Estrada, and Sabine Susstrunk.
\newblock Frequency-tuned salient region detection.
\newblock In {\em CVPR}, pages 1597--1604, 2009.

\bibitem{gcnet}
Yue Cao, Jiarui Xu, Stephen Lin, Fangyun Wei, and Han Hu.
\newblock Gcnet: Non-local networks meet squeeze-excitation networks and
  beyond.
\newblock In {\em Proceedings of the IEEE/CVF international conference on
  computer vision workshops}, pages 0--0, 2019.

\bibitem{DETR}
Nicolas Carion, Francisco Massa, Gabriel Synnaeve, Nicolas Usunier, Alexander
  Kirillov, and Sergey Zagoruyko.
\newblock End-to-end object detection with transformers.
\newblock In {\em ECCV}, pages 213--229, 2020.

\bibitem{mobileformer}
Yinpeng Chen, Xiyang Dai, Dongdong Chen, Mengchen Liu, Xiaoyi Dong, Lu Yuan,
  and Zicheng Liu.
\newblock Mobile-former: Bridging mobilenet and transformer.
\newblock In {\em Proceedings of the IEEE/CVF Conference on Computer Vision and
  Pattern Recognition}, pages 5270--5279, 2022.

\bibitem{bert}
Jacob Devlin, Ming-Wei Chang, Kenton Lee, and Kristina Toutanova.
\newblock Bert: Pre-training of deep bidirectional transformers for language
  understanding.
\newblock {\em arXiv preprint arXiv:1810.04805}, 2018.

\bibitem{vit}
Alexey Dosovitskiy, Lucas Beyer, Alexander Kolesnikov, Dirk Weissenborn,
  Xiaohua Zhai, Thomas Unterthiner, Mostafa Dehghani, Matthias Minderer, Georg
  Heigold, Sylvain Gelly, et~al.
\newblock An image is worth 16x16 words: Transformers for image recognition at
  scale.
\newblock {\em arXiv preprint arXiv:2010.11929}, 2020.

\bibitem{SSTVOS}
Brendan Duke, Abdalla Ahmed, Christian Wolf, Parham Aarabi, and Graham~W
  Taylor.
\newblock Sstvos: Sparse spatiotemporal transformers for video object
  segmentation.
\newblock In {\em CVPR}, pages 5912--5921, 2021.

\bibitem{Smeasure}
Deng-Ping Fan, Ming-Ming Cheng, Yun Liu, Tao Li, and Ali Borji.
\newblock Structure-measure: A new way to evaluate foreground maps.
\newblock In {\em ICCV}, pages 4548--4557, 2017.

\bibitem{Emeasure}
Deng-Ping Fan, Ge-Peng Ji, Xuebin Qin, and Ming-Ming Cheng.
\newblock Cognitive vision inspired object segmentation metric and loss
  function.
\newblock {\em SCIENTIA SINICA Informationis}, 6:6, 2021.

\bibitem{SSAV}
Deng-Ping Fan, Wenguan Wang, Ming-Ming Cheng, and Jianbing Shen.
\newblock Shifting more attention to video salient object detection.
\newblock In {\em CVPR}, pages 8554--8564, 2019.

\bibitem{danet}
Jun Fu, Jing Liu, Haijie Tian, Yong Li, Yongjun Bao, Zhiwei Fang, and Hanqing
  Lu.
\newblock Dual attention network for scene segmentation.
\newblock In {\em Proceedings of the IEEE/CVF conference on computer vision and
  pattern recognition}, pages 3146--3154, 2019.

\bibitem{levit}
Benjamin Graham, Alaaeldin El-Nouby, Hugo Touvron, Pierre Stock, Armand Joulin,
  Herv{\'e} J{\'e}gou, and Matthijs Douze.
\newblock Levit: a vision transformer in convnet's clothing for faster
  inference.
\newblock In {\em Proceedings of the IEEE/CVF international conference on
  computer vision}, pages 12259--12269, 2021.

\bibitem{PCSA}
Yuchao Gu, Lijuan Wang, Ziqin Wang, Yun Liu, Ming-Ming Cheng, and Shao-Ping Lu.
\newblock Pyramid constrained self-attention network for fast video salient
  object detection.
\newblock In {\em AAAI}, pages 10869--10876, 2020.

\bibitem{resnet}
Kaiming He, Xiangyu Zhang, Shaoqing Ren, and Jian Sun.
\newblock Deep residual learning for image recognition.
\newblock In {\em CVPR}, pages 770--778, 2016.

\bibitem{axialnet}
Jonathan Ho, Nal Kalchbrenner, Dirk Weissenborn, and Tim Salimans.
\newblock Axial attention in multidimensional transformers.
\newblock {\em arXiv preprint arXiv:1912.12180}, 2019.

\bibitem{flownet2}
Eddy Ilg, Nikolaus Mayer, Tonmoy Saikia, Margret Keuper, Alexey Dosovitskiy,
  and Thomas Brox.
\newblock Flownet 2.0: Evolution of optical flow estimation with deep networks.
\newblock In {\em CVPR}, pages 2462--2470, 2017.

\bibitem{FSNet}
Ge-Peng Ji, Keren Fu, Zhe Wu, Deng-Ping Fan, Jianbing Shen, and Ling Shao.
\newblock Full-duplex strategy for video object segmentation.
\newblock In {\em ICCV}, pages 4922--4933, 2021.

\bibitem{MCL}
Hansang Kim, Youngbae Kim, Jae-Young Sim, and Chang-Su Kim.
\newblock Spatiotemporal saliency detection for video sequences based on random
  walk with restart.
\newblock {\em IEEE TIP}, 24(8):2552--2564, 2015.

\bibitem{crf}
Philipp Kr{\"a}henb{\"u}hl and Vladlen Koltun.
\newblock Efficient inference in fully connected crfs with gaussian edge
  potentials.
\newblock In {\em NeurIPS}, pages 109--117, 2011.

\bibitem{EFS}
Youngjo Lee, Hongje Seong, and Euntai Kim.
\newblock Iteratively selecting an easy reference frame makes unsupervised
  video object segmentation easier.
\newblock In {\em Proceedings of the AAAI Conference on Artificial
  Intelligence}, volume~36, pages 1245--1253, 2022.

\bibitem{FGRNE}
Guanbin Li, Yuan Xie, Tianhao Wei, Keze Wang, and Liang Lin.
\newblock Flow guided recurrent neural encoder for video salient object
  detection.
\newblock In {\em CVPR}, pages 3243--3252, 2018.

\bibitem{MGA}
Haofeng Li, Guanqi Chen, Guanbin Li, and Yizhou Yu.
\newblock Motion guided attention for video salient object detection.
\newblock In {\em ICCV}, pages 7274--7283, 2019.

\bibitem{STT}
Jiangtong Li, Wentao Wang, Junjie Chen, Li Niu, Jianlou Si, Chen Qian, and
  Liqing Zhang.
\newblock Video semantic segmentation via sparse temporal transformer.
\newblock In {\em ACM MM}, pages 59--68, 2021.

\bibitem{OBN}
Siyang Li, Bryan Seybold, Alexey Vorobyov, Xuejing Lei, and C-C~Jay Kuo.
\newblock Unsupervised video object segmentation with motion-based bilateral
  networks.
\newblock In {\em Proceedings of the European conference on computer vision
  (ECCV)}, pages 207--223, 2018.

\bibitem{efficientformer}
Yanyu Li, Geng Yuan, Yang Wen, Eric Hu, Georgios Evangelidis, Sergey Tulyakov,
  Yanzhi Wang, and Jian Ren.
\newblock Efficientformer: Vision transformers at mobilenet speed.
\newblock {\em arXiv preprint arXiv:2206.01191}, 2022.

\bibitem{AFB-URR}
Yongqing Liang, Xin Li, Navid Jafari, and Jim Chen.
\newblock Video object segmentation with adaptive feature bank and
  uncertain-region refinement.
\newblock {\em Advances in Neural Information Processing Systems},
  33:3430--3441, 2020.

\bibitem{F2Net}
Daizong Liu, Dongdong Yu, Changhu Wang, and Pan Zhou.
\newblock F2net: Learning to focus on the foreground for unsupervised video
  object segmentation.
\newblock In {\em Proceedings of the AAAI Conference on Artificial
  Intelligence}, volume~35, pages 2109--2117, 2021.

\bibitem{SwinTransformer}
Ze Liu, Yutong Lin, Yue Cao, Han Hu, Yixuan Wei, Zheng Zhang, Stephen Lin, and
  Baining Guo.
\newblock Swin transformer: Hierarchical vision transformer using shifted
  windows.
\newblock In {\em ICCV}, pages 10012--10022, 2021.

\bibitem{AdamW}
Ilya Loshchilov and Frank Hutter.
\newblock Decoupled weight decay regularization.
\newblock {\em arXiv preprint arXiv:1711.05101}, 2017.

\bibitem{graphmem}
Xiankai Lu, Wenguan Wang, Martin Danelljan, Tianfei Zhou, Jianbing Shen, and
  Luc~Van Gool.
\newblock Video object segmentation with episodic graph memory networks.
\newblock In {\em European Conference on Computer Vision}, pages 661--679.
  Springer, 2020.

\bibitem{COSNet}
Xiankai Lu, Wenguan Wang, Chao Ma, Jianbing Shen, Ling Shao, and Fatih Porikli.
\newblock See more, know more: Unsupervised video object segmentation with
  co-attention siamese networks.
\newblock In {\em CVPR}, pages 3623--3632, 2019.

\bibitem{3DCSeg}
Sabarinath Mahadevan, Ali Athar, Aljo{\v{s}}a O{\v{s}}ep, Sebastian Hennen,
  Laura Leal-Taix{\'e}, and Bastian Leibe.
\newblock Making a case for 3d convolutions for object segmentation in videos.
\newblock {\em arXiv preprint arXiv:2008.11516}, 2020.

\bibitem{mobilevit}
Sachin Mehta and Mohammad Rastegari.
\newblock Mobilevit: light-weight, general-purpose, and mobile-friendly vision
  transformer.
\newblock {\em arXiv preprint arXiv:2110.02178}, 2021.

\bibitem{FBMS}
Peter Ochs, Jitendra Malik, and Thomas Brox.
\newblock Segmentation of moving objects by long term video analysis.
\newblock {\em IEEE Trans. Pattern Anal. Mach. Intell.}, 36(6):1187--1200,
  2013.

\bibitem{edgevit}
Junting Pan, Adrian Bulat, Fuwen Tan, Xiatian Zhu, Lukasz Dudziak, Hongsheng
  Li, Georgios Tzimiropoulos, and Brais Martinez.
\newblock Edgevits: Competing light-weight cnns on mobile devices with vision
  transformers.
\newblock In {\em Computer Vision--ECCV 2022: 17th European Conference, Tel
  Aviv, Israel, October 23--27, 2022, Proceedings, Part XI}, pages 294--311.
  Springer, 2022.

\bibitem{litv2}
Zizheng Pan, Jianfei Cai, and Bohan Zhuang.
\newblock Fast vision transformers with hilo attention.
\newblock {\em arXiv preprint arXiv:2205.13213}, 2022.

\bibitem{HDFNet}
Youwei Pang, Lihe Zhang, Xiaoqi Zhao, and Huchuan Lu.
\newblock Hierarchical dynamic filtering network for rgb-d salient object
  detection.
\newblock In {\em ECCV}, pages 235--252, 2020.

\bibitem{CAVER}
Youwei Pang, Xiaoqi Zhao, Lihe Zhang, and Huchuan Lu.
\newblock Caver: Cross-modal view-mixed transformer for bi-modal salient object
  detection.
\newblock {\em IEEE TIP}, 2023.

\bibitem{HFAN}
Gensheng Pei, Fumin Shen, Yazhou Yao, Guo-Sen Xie, Zhenmin Tang, and Jinhui
  Tang.
\newblock Hierarchical feature alignment network for unsupervised video object
  segmentation.
\newblock In {\em European Conference on Computer Vision}, pages 596--613.
  Springer, 2022.

\bibitem{MAE-metric}
Federico Perazzi, Philipp Kr{\"a}henb{\"u}hl, Yael Pritch, and Alexander
  Hornung.
\newblock Saliency filters: Contrast based filtering for salient region
  detection.
\newblock In {\em CVPR}, pages 733--740, 2012.

\bibitem{DAVIS}
Federico Perazzi, Jordi Pont-Tuset, Brian McWilliams, Luc Van~Gool, Markus
  Gross, and Alexander Sorkine-Hornung.
\newblock A benchmark dataset and evaluation methodology for video object
  segmentation.
\newblock In {\em CVPR}, pages 724--732, 2016.

\bibitem{youtubeobjects}
Alessandro Prest, Christian Leistner, Javier Civera, Cordelia Schmid, and
  Vittorio Ferrari.
\newblock Learning object class detectors from weakly annotated video.
\newblock In {\em 2012 IEEE Conference on computer vision and pattern
  recognition}, pages 3282--3289. IEEE, 2012.

\bibitem{RTNet}
Sucheng Ren, Wenxi Liu, Yongtuo Liu, Haoxin Chen, Guoqiang Han, and Shengfeng
  He.
\newblock Reciprocal transformations for unsupervised video object
  segmentation.
\newblock In {\em Proceedings of the IEEE/CVF conference on computer vision and
  pattern recognition}, pages 15455--15464, 2021.

\bibitem{PDB}
Hongmei Song, Wenguan Wang, Sanyuan Zhao, Jianbing Shen, and Kin-Man Lam.
\newblock Pyramid dilated deeper convlstm for video salient object detection.
\newblock In {\em ECCV}, pages 715--731, 2018.

\bibitem{segmenter}
Robin Strudel, Ricardo Garcia, Ivan Laptev, and Cordelia Schmid.
\newblock Segmenter: Transformer for semantic segmentation.
\newblock In {\em ICCV}, pages 7262--7272, 2021.

\bibitem{pwcnet}
Deqing Sun, Xiaodong Yang, Ming-Yu Liu, and Jan Kautz.
\newblock Pwc-net: Cnns for optical flow using pyramid, warping, and cost
  volume.
\newblock In {\em CVPR}, pages 8934--8943, 2018.

\bibitem{raft}
Zachary Teed and Jia Deng.
\newblock Raft: Recurrent all-pairs field transforms for optical flow.
\newblock In {\em ECCV}, pages 402--419, 2020.

\bibitem{LVOS_Visual_Memory}
Pavel Tokmakov, Karteek Alahari, and Cordelia Schmid.
\newblock Learning video object segmentation with visual memory.
\newblock In {\em Proceedings of the IEEE International Conference on Computer
  Vision}, pages 4481--4490, 2017.

\bibitem{Transformer}
Ashish Vaswani, Noam Shazeer, Niki Parmar, Jakob Uszkoreit, Llion Jones,
  Aidan~N Gomez, {\L}ukasz Kaiser, and Illia Polosukhin.
\newblock Attention is all you need.
\newblock In {\em NeurIPS}, page 5998–6008, 2017.

\bibitem{RVOS}
Carles Ventura, Miriam Bellver, Andreu Girbau, Amaia Salvador, Ferran Marques,
  and Xavier Giro-i Nieto.
\newblock Rvos: End-to-end recurrent network for video object segmentation.
\newblock In {\em Proceedings of the IEEE/CVF Conference on Computer Vision and
  Pattern Recognition}, pages 5277--5286, 2019.

\bibitem{AGNN}
Wenguan Wang, Xiankai Lu, Jianbing Shen, David~J Crandall, and Ling Shao.
\newblock Zero-shot video object segmentation via attentive graph neural
  networks.
\newblock In {\em ICCV}, pages 9236--9245, 2019.

\bibitem{AGS}
Wenguan Wang, Hongmei Song, Shuyang Zhao, Jianbing Shen, Sanyuan Zhao,
  Steven~CH Hoi, and Haibin Ling.
\newblock Learning unsupervised video object segmentation through visual
  attention.
\newblock In {\em CVPR}, pages 3064--3074, 2019.

\bibitem{pvt}
Wenhai Wang, Enze Xie, Xiang Li, Deng-Ping Fan, Kaitao Song, Ding Liang, Tong
  Lu, Ping Luo, and Ling Shao.
\newblock Pyramid vision transformer: A versatile backbone for dense prediction
  without convolutions.
\newblock In {\em ICCV}, pages 568--578, 2021.

\bibitem{pvtv2}
Wenhai Wang, Enze Xie, Xiang Li, Deng-Ping Fan, Kaitao Song, Ding Liang, Tong
  Lu, Ping Luo, and Ling Shao.
\newblock Pvt v2: Improved baselines with pyramid vision transformer.
\newblock {\em Computational Visual Media}, 8(3):415--424, 2022.

\bibitem{non-local-network}
Xiaolong Wang, Ross Girshick, Abhinav Gupta, and Kaiming He.
\newblock Non-local neural networks.
\newblock In {\em Proceedings of the IEEE conference on computer vision and
  pattern recognition}, pages 7794--7803, 2018.

\bibitem{Segformer}
Enze Xie, Wenhai Wang, Zhiding Yu, Anima Anandkumar, Jose~M Alvarez, and Ping
  Luo.
\newblock Segformer: Simple and efficient design for semantic segmentation with
  transformers.
\newblock In {\em NeurIPS}, pages 12077--12090, 2021.

\bibitem{youtubevos}
Ning Xu, Linjie Yang, Yuchen Fan, Jianchao Yang, Dingcheng Yue, Yuchen Liang,
  Brian Price, Scott Cohen, and Thomas Huang.
\newblock Youtube-vos: Sequence-to-sequence video object segmentation.
\newblock In {\em Proceedings of the European conference on computer vision
  (ECCV)}, pages 585--601, 2018.

\bibitem{RCR}
Pengxiang Yan, Guanbin Li, Yuan Xie, Zhen Li, Chuan Wang, Tianshui Chen, and
  Liang Lin.
\newblock Semi-supervised video salient object detection using pseudo-labels.
\newblock In {\em ICCV}, pages 7284--7293, 2019.

\bibitem{lvt}
Chenglin Yang, Yilin Wang, Jianming Zhang, He Zhang, Zijun Wei, Zhe Lin, and
  Alan Yuille.
\newblock Lite vision transformer with enhanced self-attention.
\newblock In {\em Proceedings of the IEEE/CVF Conference on Computer Vision and
  Pattern Recognition}, pages 11998--12008, 2022.

\bibitem{AMCNet}
Shu Yang, Lu Zhang, Jinqing Qi, Huchuan Lu, Shuo Wang, and Xiaoxing Zhang.
\newblock Learning motion-appearance co-attention for zero-shot video object
  segmentation.
\newblock In {\em ICCV}, pages 1564--1573, 2021.

\bibitem{yang2019xlnet}
Zhilin Yang, Zihang Dai, Yiming Yang, Jaime Carbonell, Russ~R Salakhutdinov,
  and Quoc~V Le.
\newblock Xlnet: Generalized autoregressive pretraining for language
  understanding.
\newblock {\em arXiv preprint arXiv:1906.08237}, 2019.

\bibitem{Anchor-Diffusion}
Zhao Yang, Qiang Wang, Luca Bertinetto, Weiming Hu, Song Bai, and Philip~HS
  Torr.
\newblock Anchor diffusion for unsupervised video object segmentation.
\newblock In {\em ICCV}, pages 931--940, 2019.

\bibitem{poolformer}
Weihao Yu, Mi Luo, Pan Zhou, Chenyang Si, Yichen Zhou, Xinchao Wang, Jiashi
  Feng, and Shuicheng Yan.
\newblock Metaformer is actually what you need for vision.
\newblock In {\em Proceedings of the IEEE/CVF conference on computer vision and
  pattern recognition}, pages 10819--10829, 2022.

\bibitem{TransportNet}
Kaihua Zhang, Zicheng Zhao, Dong Liu, Qingshan Liu, and Bo Liu.
\newblock Deep transport network for unsupervised video object segmentation.
\newblock In {\em Proceedings of the IEEE/CVF International Conference on
  Computer Vision}, pages 8781--8790, 2021.

\bibitem{DCFNet}
Miao Zhang, Jie Liu, Yifei Wang, Yongri Piao, Shunyu Yao, Wei Ji, Jingjing Li,
  Huchuan Lu, and Zhongxuan Luo.
\newblock Dynamic context-sensitive filtering network for video salient object
  detection.
\newblock In {\em ICCV}, pages 1553--1563, 2021.

\bibitem{topformer}
Wenqiang Zhang, Zilong Huang, Guozhong Luo, Tao Chen, Xinggang Wang, Wenyu Liu,
  Gang Yu, and Chunhua Shen.
\newblock Topformer: Token pyramid transformer for mobile semantic
  segmentation.
\newblock In {\em Proceedings of the IEEE/CVF Conference on Computer Vision and
  Pattern Recognition}, pages 12083--12093, 2022.

\bibitem{AMP}
Xiaoqi Zhao, Shijie Chang, Youwei Pang, Jiaxing Yang, Lihe Zhang, and Huchuan
  Lu.
\newblock Adaptive multi-source predictor for zero-shot video object
  segmentation.
\newblock {\em arXiv preprint arXiv:2303.10383}, 2023.

\bibitem{APS}
Xiaoqi Zhao, Youwei Pang, Jiaxing Yang, Lihe Zhang, and Huchuan Lu.
\newblock Multi-source fusion and automatic predictor selection for zero-shot
  video object segmentation.
\newblock In {\em ACM MM}, pages 2645--2653, 2021.

\bibitem{GateNet}
Xiaoqi Zhao, Youwei Pang, Lihe Zhang, Huchuan Lu, and Lei Zhang.
\newblock Suppress and balance: A simple gated network for salient object
  detection.
\newblock In {\em ECCV}, pages 35--51, 2020.

\bibitem{GateNetv2}
Xiaoqi Zhao, Youwei Pang, Lihe Zhang, Huchuan Lu, and Lei Zhang.
\newblock Towards diverse binary segmentation via a simple yet general gated
  network.
\newblock {\em arXiv preprint arXiv:2303.10396}, 2023.

\bibitem{setr}
Sixiao Zheng, Jiachen Lu, Hengshuang Zhao, Xiatian Zhu, Zekun Luo, Yabiao Wang,
  Yanwei Fu, Jianfeng Feng, Tao Xiang, Philip~HS Torr, et~al.
\newblock Rethinking semantic segmentation from a sequence-to-sequence
  perspective with transformers.
\newblock In {\em CVPR}, pages 6881--6890, 2021.

\bibitem{MATNet}
Tianfei Zhou, Shunzhou Wang, Yi Zhou, Yazhou Yao, Jianwu Li, and Ling Shao.
\newblock Motion-attentive transition for zero-shot video object segmentation.
\newblock In {\em AAAI}, pages 13066--13073, 2020.

\bibitem{deformable-detr}
Xizhou Zhu, Weijie Su, Lewei Lu, Bin Li, Xiaogang Wang, and Jifeng Dai.
\newblock Deformable detr: Deformable transformers for end-to-end object
  detection.
\newblock {\em arXiv preprint arXiv:2010.04159}, 2020.

\end{thebibliography}
}

\end{document}